\documentclass[10pt,twocolumn,letterpaper]{article}

\usepackage[pagenumbers]{cvpr} % To force page numbers, e.g. for an arXiv version

\usepackage{pifont}     % \ding{51} / \ding{55}
\usepackage{comment}    % \excludecomment
\usepackage{multirow}
\usepackage{colortbl}   % pulls in color + array; cell coloring in tables
\usepackage{makecell}
\usepackage{gensymb}
\usepackage{wrapfig}
\definecolor{CBRed}{RGB}{204, 102, 119}
\definecolor{CBYellow}{RGB}{221, 204, 119}
\definecolor{CBBlue}{RGB}{136, 204, 238}
\definecolor{CBGreen}{RGB}{0, 158, 115}
\definecolor{CBMagenta}{RGB}{170, 68, 153}
\definecolor{CBOrange}{RGB}{230, 159, 0}
\definecolor{tabbestcolor}{rgb}{0.785, 0.851, 0.969}

\newcommand{\OURS}{VoxLution}
\newcommand{\mypara}[1]{\vspace{3pt}\noindent\textbf{#1}~~}
\definecolor{cvprblue}{rgb}{0.21,0.49,0.74}
\usepackage[pagebackref,breaklinks,colorlinks,allcolors=cvprblue]{hyperref}

\def\confName{CVPR}
\def\confYear{2027}

\title{\OURS{}: Evolving Voxels for High-Fidelity 3D Scene Generation}

\author{%
Daoyi Gao$^{1*}$
\qquad Xinyi Zhang$^{1*}$
\qquad Alexey Bokhovkin$^{1}$
\qquad Lei Li$^{2}$
\qquad Angela Dai$^{1}$\\
$^{1}$Technical University of Munich
\qquad $^{2}$University of Virginia
}

\begin{document}

% Title + teaser figure spanning both columns of the first page.
% NOTE: figures/teaser.tex calls \maketitle inside a \twocolumn[...] block, so do
% not call \maketitle separately here. To drop the teaser, replace this \input
% with a plain \maketitle.
\twocolumn[{%
\renewcommand\twocolumn[1][]{#1}%
\maketitle
\begin{center}
  \centering
  \includegraphics[width=\textwidth]{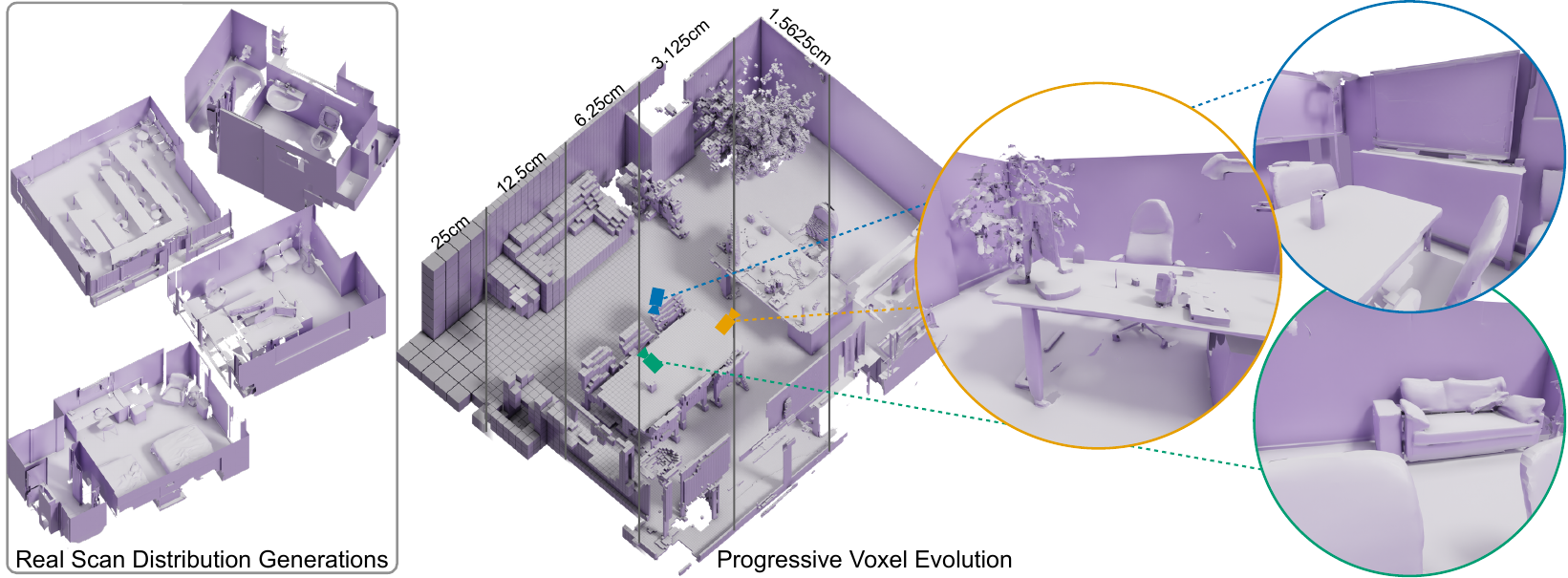}
  \captionof{figure}{%
  We introduce \OURS{}, a sparse autoregressive approach that synthesizes high-resolution 3D indoor scenes through a sparse autoregressive cascade that refines a coarse layout hierarchically, from $25\,\text{cm}$ down to $1.5625\,\text{cm}$ resolution.
  \OURS{} generates scenes that capture sharp surfaces and rich geometric details, and can be flexibly trained on synthetic or real 3D scene data; here, we show a generation from the real scan distribution.
  }
  \label{fig:teaser}
\end{center}%
\vspace{1em}
}]

% Equal-contribution note. \thanks is lost inside the teaser's \twocolumn[...] block,
% so it is set here, camera-ready only (the review version is anonymous).
\iftoggle{cvprfinal}{{\renewcommand{\thefootnote}{\fnsymbol{footnote}}\footnotetext[1]{Equal contribution.}}}{}

\begin{abstract}
\label{sec:abstract}

High-resolution 3D indoor scene generation is crucial for immersive applications such as virtual reality, spatial computing, and robotics simulation.
We introduce \OURS{}, a coarse-to-fine transformer-based approach that generates high-fidelity indoor scenes with rich details, overcoming the overly smooth geometry typical of existing generative models.
First, we generate a low-resolution spatial layout; then, based on this structural information, we progressively refine the scene geometry locally in chunks.
Key to our approach is formulating 3D super-resolution as a sequence modeling problem over multi-resolution sparse voxel grids.
To achieve this, we introduce a training-free, resolution-invariant tokenization scheme that compactly encodes the geometric information of eight octants into a single decimal number.
We formulate the transformer sequence to include multiple resolutions, allowing the coarse layout to guide the generation of fine local details.
To ensure coherence across local chunks, we incorporate adjacent chunk boundary information directly into the transformer sequencing.
\OURS{} achieves scene generation at a finer voxel resolution than existing baselines, demonstrating state-of-the-art performance on standard benchmarks {with 59.6\% lower FPD and 51.1\% lower FID on 3D-FRONT,
and 70.5\% lower FPD and 73.8\% lower FID on Scannet++}.

\end{abstract}
\section{Introduction}
\label{sec:intro}

Generating 3D indoor scenes that reflect the geometric complexity of the real world is a fundamental problem in computer graphics and vision.
Many applications in virtual reality, spatial computing, and robotics simulation require 3D environments that match the dense clutter, irregular layouts, and wide variation of object type and scale found in the real world.
For instance, immersive VR or spatial computing experiences feel sterile when populated with only tidy, sparsely furnished rooms such as those in synthetic 3D scene datasets, 
and robotics policies trained in clean synthetic apartments transfer poorly to homes with cluttered shelves or irregularly stacked objects.
Real 3D scans \cite{yeshwanth2023scannet++} capture these properties directly, yet the noise and incompleteness from sensor limits and view occlusions leave them largely untapped for generative learning.

Generating realistic 3D scenes is intrinsically challenging.
Scenes vary enormously in spatial extent, with no standard canonicalization analogous to the single-object setting.
A good generator must not only capture coherent global structure in scene layouts, but also produce fine-scale detail across furniture and object clutter. 
Faithfully representing such detail requires high spatial resolution, which can impose significant demands due to the high dimensionality of 3D space, particularly for fine-grained resolutions where the surface represents a small percentage of a full 3D volume.

As real 3D scenes amplify many of these challenges, work on 3D scene generation focuses on training on synthetic 3D scene data \cite{fu20213d}, which can simplify some of these constraints by focusing on complete, axis-aligned layouts with limited clutter.
Building on this foundation, recent generative approaches have demonstrated remarkable progress~\cite{wu2024blockfusion, meng2025lt3sd,li2026worldgrow, meng2026seen2scene}, producing increasingly compelling and structurally coherent environments by operating over compressed latent grids that allow higher resolution than dense voxel representations.

In contrast, we introduce \OURS{}\footnote{Project page: \href{https://daoyig.github.io/VoxLution/}{daoyig.github.io/VoxLution/}}, 
an autoregressive generative model over discrete scene geometry tokens designed to match the irregular structure of scene geometry in 3D space.
\OURS{} operates over a sparse scene hierarchy, built on a shared, resolution-invariant voxel codebook.
Central to our approach is a training-free tokenization that generalizes across spatial resolutions: each voxel's eight child octants are encoded as a single byte, yielding a fixed 256-entry codebook shared across all resolution levels.
We then serialize the multi-resolution scene hierarchy into a single token sequence ordered from coarsest to finest, with coarser levels directly conditioning the generation of finer details, casting scene synthesis as a coarse-to-fine super-resolution task.
Our approach natively supports the inherent sparsity and incompleteness of real 3D scene scans,
generating tokens only where geometry exists, 
adaptively omitting empty or unobserved regions,
and thus enabling scalable modeling of fine-grained geometry details.

Concretely, the hierarchical generation starts from a coarse scene structure and refines geometry chunk by chunk, doubling resolution at each stage.
Each new chunk generation is conditioned on boundary voxels of already-generated neighbors, ensuring coherent chunk transitions.
A mesh extractor finally lifts the scene voxels into a triangle mesh by predicting per-voxel dual-contouring attributes.
Because the voxel codebook and sparse hierarchy formulation are dataset-agnostic, \OURS{} can be easily adapted and trained directly on real-world scans, capturing their irregular, sparse, and noisy geometry.

Experiments on 3D indoor scene datasets show that \OURS{} synthesizes scenes at a {41\% finer voxel resolution than existing baselines, while improving geometric and visual quality, with 59.6\% lower FPD and 51.1\% lower FID than the baselines on 3D-FRONT.}
{On real-world scans, \OURS{} generates high-fidelity scenes that capture the irregularity and clutter of real indoor environments, achieving 70.5\% lower FPD and 73.8\% lower FID than the baselines on Scannet++.}

In summary, our contributions are:
\begin{itemize}[leftmargin=*, labelwidth=1.5em, labelsep=0.5em, topsep=0em, itemsep=0em, parsep=0em]
    \item We propose \OURS{}, a sparse autoregressive approach that synthesizes high-fidelity 3D indoor scenes on multi-resolution voxel octrees, exploiting scene-level sparsity to scale hierarchical generation to real-world resolutions.
    \item We introduce a resolution-invariant octant code (ROC), a training-free tokenization of voxel octants with a single discrete codebook shared across all resolution levels.
    \item Building on ROC, we design sparse multi-resolution token sequences with spatially-grounded boundary condition, enabling seamless chunk-by-chunk super-resolution.
    \item We adapt \OURS{} to real-world data, extending high-resolution scene generation to noisy, incomplete indoor scans.
\end{itemize}

\section{Related Works}
\label{sec:related_works}

\subsection{3D Indoor Scene Generation}

Approaches to 3D indoor scene generation can be broadly grouped into three families: direct geometric synthesis, layout-based composition with retrieved assets, and lifting of 2D generative priors into 3D. 
Direct geometric methods generate scene-scale geometry from scratch, typically using diffusion-based approaches over compressed representations. 
BlockFusion~\cite{wu2024blockfusion}, LT3SD~\cite{meng2025lt3sd}, and WorldGrow~\cite{li2026worldgrow} use coarse-to-fine paradigms over per-block triplanes, latent trees of TUDF and feature grids, and structured latents inherited from an object-centric foundation model~\cite{xiang2024trellis}. Other works explore neural-field latents~\cite{bautista2022gaudi,
kim2023neuralfieldldm,chen2023scenedreamer}, sparse-voxel diffusion~\cite{ju2024diffindscene,ren2024xcube}, and compositional triplanes~\cite{yan2024frankenstein}. 
In contrast, \OURS{} operates directly on sparse voxel occupancy and uses an autoregressive transformer to achieve finer resolution without learned latent compression. 

Layout-based methods instead cast scene synthesis as predicting an arrangement of object bounding boxes whose geometry is retrieved from a CAD library~\cite{ritchie2019fast,wang2021sceneformer,paschalidou2021atiss,tang2024diffuscene,lin2024instructscene,wei2023legonet,yang2024physcene,zhai2023commonscenes,zhai2024echoscene}, 
restricting output to the catalog and excluding novel objects and fine-scale clutter. 

Finally, a complementary group of methods lifts pretrained 2D image and video diffusion priors to 3D using depth estimation and iterative inpainting~\cite{hollein2023text2room,fridman2023scenescape,schult2024controlroom3d,chung2023luciddreamer,yu2024wonderjourney,yu2025wonderworld}, achieving photorealistic appearance but suffering from geometric drift and view inconsistencies.

\subsection{Autoregressive Models for 3D Geometry}

Autoregressive transformers \cite{vaswani2017attention} have emerged as a strong paradigm for 3D shape generation, reflecting their dominance in language and image modeling. 
Mesh-based methods generate polygonal geometry as token sequences, from the seminal work PolyGen~\cite{nash2020polygen} to modern MeshGPT~\cite{siddiqui2024meshgpt} and its successors~\cite{chen2024meshxl,chen2024meshanything,gao2025meshart}. Beyond meshes, AR has been applied to other representations. AutoSDF~\cite{mittal2022autosdf} autoregresses over a latent SDF grid, whereas other methods cover sparse occupancy~\cite{yan2022shapeformer}, point clouds~\cite{sun2020pointgrow,chen2023pointgpt}, and octree-based representation~\cite{wang2023octformer,wei2025octgpt}. At the scene scale, AR has been applied only at the layout abstraction -- ATISS~\cite{paschalidou2021atiss} and SceneFormer~\cite{wang2021sceneformer} predict ordered or unordered sequences of object-attribute tokens rather than scene geometry. The recent work in the image domain, VAR~\cite{tian2024var}, has introduced multi-scale autoregressive modeling via next-scale prediction over hierarchical tokens, achieving quality competitive with diffusion transformers at faster inference, followed successors~\cite{chen2025sar3d,zhang2025g3pt}. \OURS{} extends multi-scale AR prediction to 3D scene geometry using sparse voxel hierarchies and a {training-free,} resolution-invariant tokenization.

\subsection{Hierarchical and Sparse 3D Representations}

Sparse hierarchical data structures and surface extraction methods have undergone a long development in 3D geometry processing. 
Octree-based representations such as OctNet~\cite{riegler2017octnet} and O-CNN~\cite{wang2017ocnn,wang2018adaptiveocnn}, alongside sparse 3D convolution approaches~\cite{choy2019minkowski,graham2018sparseconvnet}, established hierarchical sparse voxels as efficient alternatives to dense grids. 
Subsequent methods extend these structures to coarse-to-fine generative modeling, including hierarchical neural implicit functions in OctField~\cite{tang2022octfield} and NKSR~\cite{huang2023nksr}, sparse-voxel latent diffusion~\cite{ren2024xcube}, and multi-scale generative methods learned from single shapes~\cite{wei2024sin3dm,hu2023sin3dgen,hane2017hsp}. Surface extraction from such representations has also significantly developed, from classical Marching Cubes~\cite{lorensen1987marching} and Dual Contouring~\cite{ju2002dual} to recent neural variants such as FlexiCubes~\cite{shen2023flexicubes}, which learn per-voxel vertex offsets and edge attributes for sharper geometry~\cite{shen2021dmtet,chen2021neuralmc}. \OURS{} operates on a sparse multi-resolution voxel octree and extracts sharp meshes using the Flexible Dual Grid representation~\cite{xiang2025native}.

\section{Method}
\label{sec:method}

\OURS{} synthesizes 3D indoor scenes as multi-resolution voxel octrees, in an autoregressive, coarse-to-fine fashion conditioned on scene-type labels. 
As 3D surfaces are surrounded by empty space, and this ratio worsens dramatically at finer resolutions, both our tokenization and sequencing are designed around sparsity.

To compress local geometry, our \emph{Resolution-Invariant Octant Coding (ROC)} (\cref{subsec:tokenization}) maps each parent voxel's 8-child octant occupancy to a single discrete token, from a codebook shared across all resolutions.
To compress the sequence itself, sparse multi-resolution sequences (\cref{subsec:sequencing}) drop empty tokens and pack hierarchical context and spatial neighbor information into a 1D sequence.
A family of autoregressive predictors $\{\Phi^{(k)}\}$ is then used to synthesize a scene as a multi-resolution hierarchy: $\Phi^{(0)}$ generates a coarse scene structure $O^{(0)}$, and $\Phi^{(k)}$ refines it chunk by chunk at increasing resolutions, yielding a sparse occupancy $O^{(K)}$ at the finest voxel resolution (\cref{subsec:generation}).
Finally, a sparse VAE $\Phi_m$ (\cref{subsec:mesh}) lifts $O^{(K)}$ into an output triangle mesh $\mathcal{M}$.
As our method adopts occupancy-based generation with a final mesh extraction, it can be flexibly applied to both synthetic and real scan data.
Beyond scene-type labels, \OURS{} supports layout-controlled generation, conditioning every stage on a semantic layout of object bounding boxes (\cref{subsec:layout}).

\begin{figure*}[t]
  \centering
  \includegraphics[width=\textwidth]{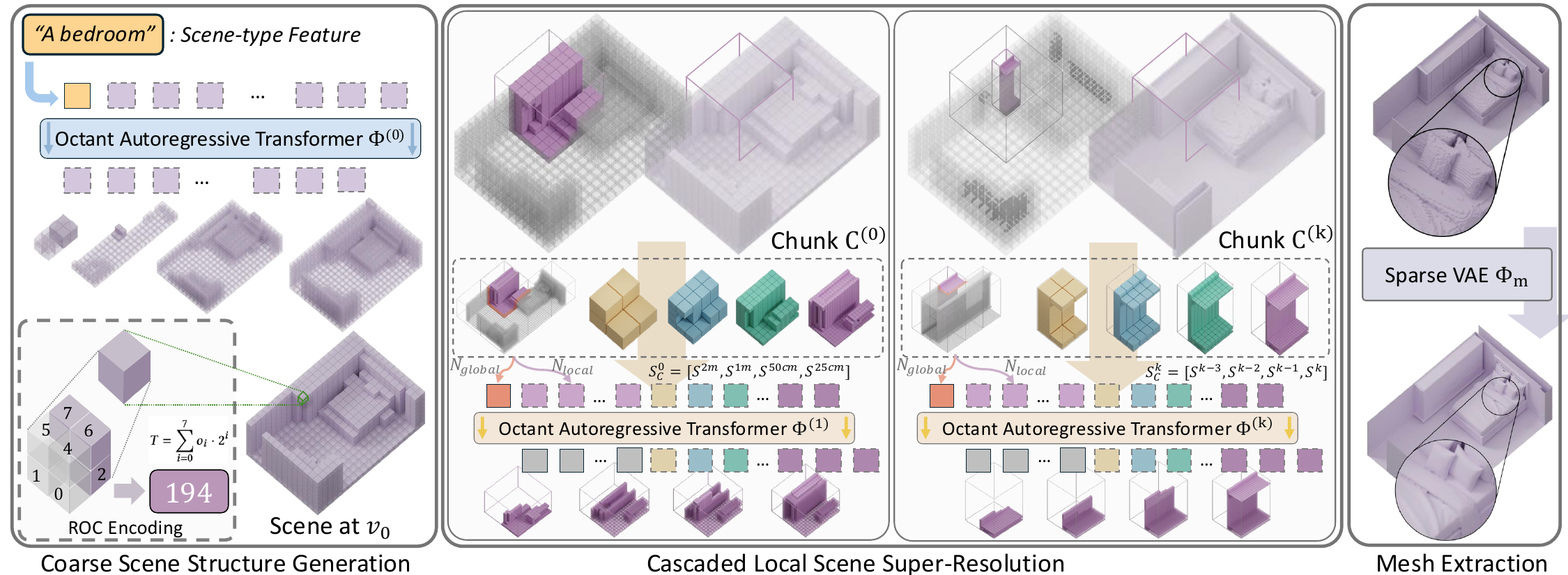}
  \caption{%
    Overview of \OURS{}.
    We synthesize high-fidelity 3D indoor scenes as a multi-resolution autoregressive cascade, using a sparse resolution-invariant coding (ROC) for tokenization and chunked multi-resolution sequencing with cross-level and boundary neighbor conditioning.
    \textbf{Left}: 
    {$\Phi^{(0)}$ generates the initial coarse scene structure as a single-level ROC sequence conditioned on the input scene-type label.}
    \textbf{Middle}: each cascade transformer $\Phi^{(k)}$ ($k \geq 1$) consumes a sparse multi-resolution sequence with coarser-level and neighboring context, refining chunks to finer voxel resolution.
    \textbf{Right}: a sparse VAE $\Phi_m$ decodes the final voxel grid into  dual-contouring attributes that produce the final triangle mesh.%
  }
  \label{fig:pipeline}
\end{figure*}

\subsection{Resolution-Invariant Octant Coding}
\label{subsec:tokenization}

To leverage the sequence-modeling capabilities of transformers, 3D data needs mapping into a discrete codebook.
Rather than relying on learned codebooks~\cite{van2017neural}, which typically requires resolution-specific training, we instead introduce a \emph{resolution-invariant octant code} (ROC), a training-free encoding that compactly represents local 3D geometry with  a single codebook shared across all resolution levels.
 
Given a parent voxel $V$ at resolution level $k$, we uniformly subdivide it into 8 equal-sized child voxels (octants) at the next finer level $k{+}1$.
Let $o_i \in \{0, 1\}$ denote the binary occupancy of the $i$-th child under a $z$--$y$--$x$ traversal order ($i \in \{0, \dots, 7\}$).
The 8-child occupancy pattern of $V$ then takes one of $2^8 = 256$ possible patterns, each of which we treat as a codebook entry. The token $T \in \{0, \dots, 255\}$ directly encodes this as
\begin{equation}
    T = \sum_{i=0}^{7} o_i \cdot 2^i.
\end{equation}

\mypara{Compactness and resolution-invariance.}
ROC compresses geometry for our autoregressive scene hierarchy generation.
The eight child voxels of each parent are collapsed into a single token, reducing the spatial sequence length $8\times$ and substantially relieving the attention computation bottleneck.
ROC is also resolution-invariant by construction: the encoding $T$ depends only on the local 8-octant occupancy pattern, so the same 256-entry codebook is
valid at every level of the octree and a code carries the same geometric meaning regardless of the absolute voxel size -- $T{=}240$ always denotes the upper-$z$ slab whether
the octants are $1\,\text{m}$ or $1.5625\,\text{cm}$ cubes.
In contrast, learned tokenizers such as VQ-VAEs~\cite{van2017neural} tie the codebook to a specific resolution and require a separate quantizer per scale.
A single shared codebook is thus reused across all cascade transformers without retraining.

\subsection{Sparse Multi-Resolution Token Sequencing}
\label{subsec:sequencing}
Autoregressive prediction over voxel octrees requires serializing hierarchical occupancy into a 1D token sequence. At fine voxel resolutions a full-scene sequence becomes too long for a single transformer pass, so we partition the scene into non-overlapping axis-aligned chunks and serialize one chunk at a time. 
The serialization orders ROC tokens within each resolution, stitches together different resolutions of the same chunk, and brings in context from already-generated neighboring chunks, with a disentangled positional encoding identifying the role of each token.

\mypara{Intra-resolution sparse sequencing.}
We index resolution levels by $k$, with voxel size $v_k = v_0 / 2^{k}$, where $v_0 = 25\,\text{cm}$ is the voxel size of the coarse structure $O^{(0)}$; levels $k<0$ (\eg, $v_{-1} = 50\,\text{cm}$) are obtained by $2{\times}2{\times}2$ max-pooling.
We refer to the ROC tokens of the occupied level-$(k{-}1)$ voxels as \emph{level-$k$ tokens}, as their octants are the level-$k$ voxels, \ie, they encode the occupancy $O^{(k)}$.
Level-$k$ tokens inherit a recursive $z$--$y$--$x$ order from the octree: each parent's eight child tokens appear contiguously in $z$--$y$--$x$ order, and the parents themselves are listed in the same recursive order at the level above.
Equivalently, level-$k$ tokens follow the Morton (Z-order) curve along $z$--$y$--$x$~\cite{morton1966computer}.
As indoor scenes contain mostly empty space, we omit $T{=}0$ tokens, yielding {for each chunk $C$ (defined below) a sparse level-$k$ token sequence $S_C^{(k)}$} that scales with occupied geometry rather than volume.

\mypara{Scene chunk partitioning.}
\label{para:chunking}
At each level $k$, the scene is divided into a regular grid of non-overlapping axis-aligned chunks $\{C\}$, each covering a fixed {square} footprint at the floor plane and spanning the full scene height. 
This keeps per-input sequence length tractable at the scale of 3D scenes, where occupancy grows cubically with resolution.

A chunk $C$ carries the occupancy of its region across all resolution levels available so far, so $C$'s tokens come from multiple resolutions of the same volume.
From one cascade level to the next, each chunk is split into $2{\times}2$ children in the floor plane while the voxel size is halved. At a given level, chunks are generated in raster-scan order, so when $C$ is processed its left, back, and corner neighbors $\mathcal{N}(C)$ are already finished and available as context.

\mypara{Coarse-to-fine resolution conditioning.}
Within a chunk $C$, generating the level-$k$ tokens relies on the coarser levels of $C$.
Our multi-resolution token sequence {$\mathcal{S}_C^{(k)}$} places the three immediately coarser levels as conditioning context for the target level sequence, concatenated coarse to fine,
\begin{equation}
    \mathcal{S}_C^{(k)} = \big[\, S_C^{(k-3)},\; S_C^{(k-2)},\; S_C^{(k-1)},\; S_C^{(k)}\,\big],
\end{equation}
separated by a learned resolution-level split token that marks each transition between consecutive levels.
Bounding the context to three coarser levels keeps the per-chunk sequence length manageable while giving each fine prediction direct access to the surrounding structure at multiple scales. At early cascade stages where fewer than three coarser levels arise naturally from the preceding stages, we {obtain the missing levels $k<0$ by max-pooling $O^{(0)}$}.

\mypara{Boundary conditioning.}
Chunks within a cascade resolution are generated in raster-scan order, and must align with their previously generated neighbors at shared boundaries.
The previously generated left, back, and corner neighbors $\mathcal{N}(C)$ are used as context to condition the generation of $C$.

To provide holistic context, we form a global neighbor summary $\mathcal{B}_{\text{global}}$ by compressing the full multi-resolution sequence of each $C' \in \mathcal{N}(C)$ into a single latent vector via a shared GRU~\cite{dey2017gate}, prepended to {$\mathcal{S}_C^{(k)}$}.

For voxel-level boundary alignment, we form a spatial 1-ring prefix $\mathcal{B}_{\text{local}}$ from the tokens of each $C' \in \mathcal{N}(C)$ that lie on its shared boundary with $C$, at every available resolution. Each prefix token keeps its 3D position within $C'$, so the model sees boundary voxels at their actual spatial locations rather than at a generic prefix slot.
The generation of the target level is then:
\begin{equation}
    {p\!\left(S_C^{(k)} \,\Big|\, \mathcal{B}_C, S_C^{(<k)} \right) \;=\; \prod_{j=1}^{|S_C^{(k)}|} p\!\left(s_j \,\Big|\, \mathcal{B}_C, S_C^{(<k)}, s_{<j}\right),}
\end{equation}
where {$s_j$ is the $j$-th token of $S_C^{(k)}$, $\mathcal{B}_C = (\mathcal{B}_{\text{global}}, \mathcal{B}_{\text{local}})$}
denotes the boundary context {of $C$}, and {$S_C^{(<k)} = [S_C^{(k-3)}, S_C^{(k-2)}, S_C^{(k-1)}]$} denotes the coarser conditions.

\mypara{Cascade sequence positional encoding.}
\label{para:disentangled-pe}
A token in a cascade chunk sequence {$\mathcal{S}_C^{(k)}$ carries} several distinct attributes, each with different characteristics: hierarchical (resolution level), spatial (3D position within {its chunk}), and structural (role as boundary token from a specific neighbor direction or {a token of $C$}).
We encode each attribute with learned embeddings and sum them:
\begin{equation}
    {\mathrm{PE}(s) \;=\; E_{\text{hier}}(\ell_s) \;+\; E_{\text{spatial}}(\mathbf{p}_s) \;+\; E_{\text{type}}(t_s),}
\end{equation}
where {$E_{\text{hier}}$, $E_{\text{spatial}}$, and $E_{\text{type}}$ are the learned embedding tables, $\ell_s$} is the resolution level of token $s$, $\mathbf{p}_s = (x,y,z)$ is {its local 3D coordinate within its chunk ($C$, or $C'$ for a boundary token from neighbor $C'$), with $E_{\text{spatial}}(\mathbf{p}_s) = E_x(x) + E_y(y) + E_z(z)$ decomposed across axes}, and $t_s$ is a discrete type identifying whether the token is a boundary token from a specific neighbor direction or {a token of $C$.
$\mathrm{PE}(s)$ is added to the token embedding of $s$ in place of the standard 1D positional embedding.}

\subsection{Autoregressive Scene Generation}
\label{subsec:generation}

Based on the sparse multi-resolution sequencing of Sec.~\ref{subsec:sequencing}, we can now formulate autoregressive scene generation as sampling chunk sequences and stitching them into a full scene. A coarse predictor $\Phi^{(0)}$ and $K$ refinement
predictors $\{\Phi^{(k)}\}_{k=1}^{K}$ are all decoder-only transformers sharing the same architecture.
Each predictor is trained by next-token prediction on voxelized occupancy data, which can come from either synthetic scenes or real-world scans without changes to the architecture. 

The initial generator $\Phi^{(0)}$ consumes {a single-level dense sequence without coarser levels and is conditioned on a prepended scene-type label}, while $\Phi^{(k)}$ for $k \geq 1$ consumes sparse multi-resolution sequences {$\mathcal{S}_C^{(k)}$} with three coarser conditioning levels and is conditioned on the boundary context $\mathcal{B}_C$ of already-generated neighbors.

\mypara{Coarse scene generation.}
Generation begins with $\Phi^{(0)}$ producing the coarse structure of a scene.
This can be considered a degenerate case of the multi-resolution sequencing: {there are no coarser levels to condition on, and the coarse sequence is short enough to be serialized directly as a dense sequence.}
The scene is partitioned into coarse chunks at $v_{0} = 25\,\text{cm}${,} which are generated in raster-scan order.
{Each coarse chunk is encoded by its dense level-$0$ tokens $S_C^{(0)}$, \ie, one ROC token per $50\,\text{cm}$ voxel with $T{=}0$ tokens kept, and we serialize all chunks into a single 1D dense sequence, so that each chunk attends to all previously generated chunks.}
We condition on a scene-type label ({\eg,} bedroom, living room, office), encoded by a frozen text encoder \cite{radford2021learning} and linearly projected into the transformer feature space.
The generated chunks then collectively compose the coarse occupancy grid $O^{(0)}$.

\mypara{Cascaded chunk refinement.}
\label{para:chunk_refinement}
Given $O^{(0)}$, a cascade of transformers $\Phi^{(k)}$ then synthesizes geometric details at successively finer resolutions to produce occupancies $O^{(k)}$.
Each $\Phi^{(k)}$ conditions on the coarser outputs $O^{(k-3)}, \dots, O^{(k-1)}$ already generated. 
As resolution doubles, we shrink the chunk footprint accordingly to maintain manageable per-chunk sequence lengths.
Note that $k\geq 1$ cascade sequences are sparse with $T{=}0$ tokens dropped, mirroring the inherent sparsity of 3D surface geometry.

Chunks are processed in raster-scan order at each resolution level $k$. For each chunk, $\Phi^{(k)}$ reads tokens at three coarser resolutions from the prior outputs at the chunk's spatial extent, 
packs them with the target level into {$\mathcal{S}_C^{(k)}$}, and autoregressively generates the chunk's {level-$k$ tokens $S_C^{(k)}$}, which populate $O^{(k)}$. The cascade ends after $K$ stages with the finest occupancy $O^{(K)}$ at voxel size $v_K$, which is passed to the mesh decoder.

\begin{figure*}[t]
  \centering
  \includegraphics[width=\textwidth]{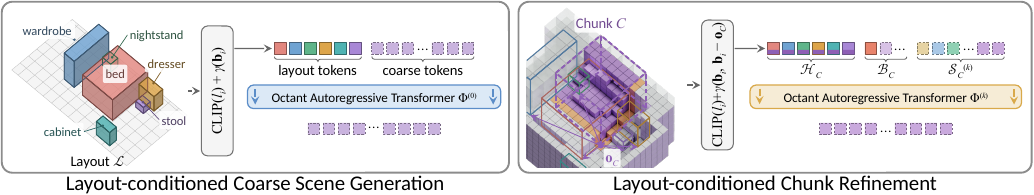}
  \caption{%
    {Layout-conditioned generation.
    \textbf{Left}: We condition the generation on the layout by embedding each box $(\mathbf{b}_i, l_i)$ of the layout $\mathcal{L}$ as a layout token.
    The layout tokens replace the scene-type label as the prefix of the coarse sequence of $\Phi^{(0)}$.
    \textbf{Right}: for each chunk $C$, the layout tokens $\mathcal{H}_C$ additionally encode each box relative to the chunk origin $\mathbf{o}_C$, and precede the boundary context $\mathcal{B}_C$ (orange: 1-ring neighbor boundary) and the multi-resolution sequence $\mathcal{S}_C^{(k)}$ of $\Phi^{(k)}$.}%
  }
  \label{fig:layout_pipeline}
\end{figure*}

\subsection{{Layout-Controlled Scene Generation}}
\label{subsec:layout}

{A scene-type label specifies only the kind of room to generate, whereas many applications also require control over which objects a scene contains and where they are placed, \eg, interior design, where furniture is arranged within a given room, and simulation environments for embodied agents, where task-relevant objects must appear at specified locations.
Following Seen2Scene~\cite{meng2026seen2scene}, we provide this control through a \emph{semantic layout}, a set of object bounding boxes with semantic labels.
The layout specifies only the category and placement of each object, while \OURS{} generates the remaining geometry: the fine detail of each object, the room structure such as walls and floors.}

\mypara{{Layout representation.}}
{We represent layout as a set of axis-aligned boxes with semantic labels, $\mathcal{L} = \{(\mathbf{b}_i, l_i)\}_{i=1}^{N}$, where $\mathbf{b}_i \in \mathbb{R}^6$ holds the box corners normalized by the room extent.
Each box is embedded as a \emph{layout token} $\mathbf{h}_i = W_l\,\mathrm{CLIP}(l_i) + W_b\,\gamma(\mathbf{b}_i)$, the sum of a projection of the CLIP text feature~\cite{radford2021learning} of its label $l_i$ and a projection of the Fourier features $\gamma(\cdot)$~\cite{tancik2020fourier} of its continuous coordinates $\mathbf{b}_i$.}
% The open-vocabulary label embedding enables joint training on datasets with different annotation vocabularies.}

\mypara{{Layout-conditioned coarse structure generation.}}
{$\Phi^{(0)}$ receives the layout through two complementary channels.
The layout tokens replace the scene-type label as the prefix of the coarse sequence (\cref{fig:layout_pipeline}, left) and carry the exact box extents, while a semantic field added onto the coarse token grid informs each token which objects occupy its cell.
For each $50\,\text{cm}$ cell, the field is the overlap-weighted mean of the CLIP features of the boxes covering the cell.}

\mypara{{Layout-conditioned chunk refinement.}}
{Each $\Phi^{(k)}$ takes the layout tokens as a prefix: the tokens $\mathcal{H}_C$ of the boxes nearest to $C$ are inserted ahead of the boundary context (\cref{fig:layout_pipeline}, right), and the target level is generated as $p\big(S_C^{(k)} \,|\, \mathcal{H}_C, \mathcal{B}_C, S_C^{(<k)}\big)$.
{Each token in $\mathcal{H}_C$ additionally encodes its box relative to the chunk origin $\mathbf{o}_C$, the minimum corner of $C$, as $W_l\,\mathrm{CLIP}(l_i) + W_b'\,\gamma(\mathbf{b}_i, \mathbf{b}_i - \mathbf{o}_C)$, which relates the layout to the voxels of $C$.}
We refer the reader to \cref{supp:layout_guided} for more details.}

\subsection{Mesh Extraction}
\label{subsec:mesh}

The cascade produces a sparse occupancy grid at the finest voxel size $v_K$, but downstream rendering and simulation often require a mesh.
Naive marching-cubes extraction~\cite{lorensen1998marching} from binary occupancy yields blocky, axis-aligned artifacts rather than sharp details that match the finest {$v_K = 1.5625\,\text{cm}$ voxel size} we achieve.
We instead extract the mesh from this sparse occupancy with a sparse VAE $\Phi_m$.

We adopt the Flexible Dual Grid representation~\cite{xiang2025native}, which describes a surface by associating each occupied voxel with a dual-vertex offset within the cell and a 3-bit edge-intersection mask along the positive axes. 
A single dual-contouring pass~\cite{ju2002dual} over these attributes yields output mesh $\mathcal{M}$, with quad faces split into triangles using a tessellation-time quad-lerp weight.

\mypara{{Generative sparse VAE.}}
{The encoder takes the voxel occupancy $O^{(K)}$ as input and progressively downsamples it into a sparse latent.
Unlike the original SC-VAE~\cite{xiang2025native}, whose decoder reproduces the input voxel set of the encoder, our decoder predicts the child occupancy at every upsampling level, so that it can both add and remove voxels.
This enables $\Phi_m$ to fill small holes and remove stray voxels in the generated occupancy.}

\mypara{{Training the generative sparse VAE.}}
{We train $\Phi_m$ with a per-level binary cross-entropy on child occupancy, losses on the dual-grid attributes, a KL regularizer, and the rendering losses of TRELLIS.2~\cite{xiang2025native}, using cameras placed inside each room.
During training, we augment the input with synthetic occupancy artifacts, such as missing voxel patches, while supervising the decoder with the clean ground truth.
We refer the reader to \cref{supp:mesh} for the details.}

\section{Experiments}
\label{sec:experiments}

We evaluate \OURS{} on 3D indoor scene generation on both synthetic scenes from 3D-FRONT~\cite{fu20213d} and real-world scans from ScanNet++~\cite{yeshwanth2023scannet++}, comparing against the state of the art on geometric fidelity and visual quality.

\mypara{Datasets.}
We train and evaluate \OURS{} on two indoor scene datasets: 3D-FRONT~\cite{fu20213d} for synthetic scenes and ScanNet++~\cite{yeshwanth2023scannet++} for real-world scans.
3D-FRONT is a large-scale collection of synthetic indoor scenes with diverse layouts and furniture arrangements.
We use five room categories, including living room, bedroom, living-dining room, library, and kids room, yielding 19{,}237 rooms.
From ScanNet++, we filter out rooms whose scans contain large unscanned regions, leaving 781 rooms.
Each dataset is split 80\%/5\%/15\% into train/validation/test, and \OURS{} and all baselines use the same splits.

\begin{figure*}[hbt]
  \centering
  \includegraphics[width=\textwidth,trim={0 20 0 0},clip]{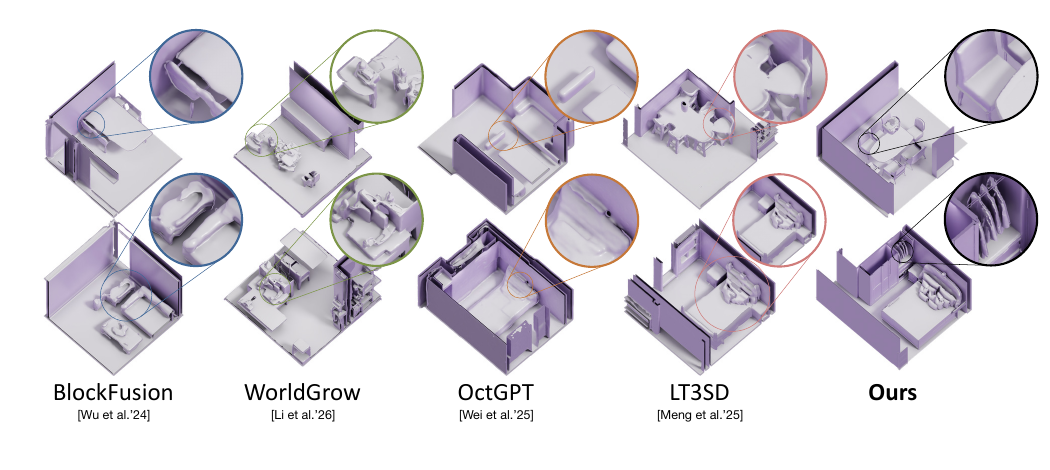}
  \par\vspace{-6pt}
  \caption{%
    Qualitative comparison on 3D-FRONT~\cite{fu20213d}. \OURS{} generates sharp surfaces and rich geometric details.
  }
  \label{fig:qualitative_comparison}
\end{figure*}

\mypara{Implementation details.}
{We implement \OURS{} in PyTorch. The coarse scene transformer $\Phi^{(0)}$ and the $K{=}4$ cascade transformers share a GPT-2 decoder backbone with 24 layers, a hidden dimension of 1024, and 16 attention heads; each trains for $\approx 4$ days on four NVIDIA H100 GPUs.}
{The mesh VAE $\Phi_m$ is initialized from the pretrained dual-grid SC-VAE~\cite{xiang2025native} and trained on 3D-FRONT, ScanNet++, and ARKitScenes~\cite{baruch2021arkitscenes}.}
{More details in \cref{supp:impl}.}

\mypara{Baselines.}
We compare with state-of-the-art 3D scene generation methods: {WorldGrow}~\cite{li2026worldgrow}, {BlockFusion}~\cite{wu2024blockfusion}, and {LT3SD}~\cite{meng2025lt3sd}, all designed for synthetic scene training. 
{In addition, we adapted OctGPT~\cite{wei2025octgpt} from object- to scene-level generation by fine-tuning both its VQ-VAE and transformer on scene datasets.}
{For layout-controlled scene generation, we compare with Seen2Scene~\cite{meng2026seen2scene}, which generates scenes conditioned on a semantic layout of object bounding boxes.}

\mypara{Evaluation metrics.}
We measure scene quality with distribution distances at two levels.
At the geometric level, we report the {\textit{Fr\'echet Point Distance (FPD)}~\cite{zhang20233dshape2vecset}} between point clouds sampled from generated and ground-truth meshes{, computed on the features of the pretrained point cloud encoder Uni3D~\cite{zhou2024uni3d}}.
At the visual level, we render each scene and report the \textit{Fr\'echet Inception Distance (FID)}~\cite{heusel2017gans} and the \textit{Kernel Inception Distance (KID)}~\cite{Binkowski2018KID} against ground-truth renderings{, computed on DINOv2 ViT-B/14 features~\cite{oquab2023dinov2} in place of Inception features}. Lower is better for all three.
{See \cref{supp:eval_protocol} for the details of the protocol.}

\subsection{Comparison to State of the Art}

% Table 1 with the updated metrics (FPD on normalized point clouds, DINOv2-B FID / KID).
% Ours = open_longitude. The previous version is tables/main_quantitatives.tex.
\begin{table}[t]
    \centering
    \caption{Quantitative evaluation on synthetic 3D scene generation. Our approach outperforms baselines in both geometric and perceptual metrics. {WorldGrow, BlockFusion, and LT3SD are evaluated with their released checkpoints, which are trained on 3D-FRONT;} {$^{*}$~denotes finetuning the released checkpoint on 3D-FRONT.}
    % \todo{mention others were originally trained on 3df}
    }
    \label{tab:main_results}
    \small
    \setlength{\tabcolsep}{4pt}
    \begin{tabular}{lccc}
        \toprule
        Method & {FPD$\times 10^{2}$} ($\downarrow$) & {FID$\times 10^{-2}$} ($\downarrow$) & KID$\times 10^{2}$ ($\downarrow$) \\
        \midrule
        WorldGrow~\cite{li2026worldgrow} & {41.85} & {8.49} & {12.52} \\
        BlockFusion~\cite{wu2024blockfusion} & {56.54} & {5.57} & {4.13} \\
        {OctGPT$^{*}$~\cite{wei2025octgpt} } & {37.76} & {8.24} & {8.11} \\
        LT3SD~\cite{meng2025lt3sd} & {27.31} & {4.91} & {3.98} \\
        \midrule
        % NOTE:  must be applied per cell -- a macro argument cannot span "&"
        {\textbf{\OURS{} (Ours)}} & {\textbf{11.03}} & {\textbf{2.40}} & {\textbf{1.49}} \\
        \bottomrule
    \end{tabular}
\end{table}

% Table 2 with the updated metrics (FPD on normalized point clouds, DINOv2-B FID / KID).
% Ours = genL_mix3_0902_merged4. The previous version is tables/scannetpp_results.tex.
\begin{table}[t]
    \centering
    \caption{Quantitative evaluation on real-world ScanNet++ dataset. Our approach outperforms baselines in both geometry fidelity and perceptual realism. $^{*}$~denotes finetuning the released checkpoint on {our ScanNet++ training split}.
    % \ANGIE{we need to show the layout vs not layout conditioned results here and not stuff them into the supp, we are using those for the main visuals}
    }
    \label{tab:scannetpp_results}
    \small
    \setlength{\tabcolsep}{4pt}
    \begin{tabular}{lccc}
        \toprule
        Method & {FPD$\times 10^{2}$} ($\downarrow$) & {FID$\times 10^{-2}$} ($\downarrow$) & KID$\times 10^{2}$ ($\downarrow$) \\
        \midrule
        {OctGPT$^{*}$~\cite{wei2025octgpt}} & {46.14} & {10.21} & {10.45} \\
        LT3SD$^{*}$~\cite{meng2025lt3sd} & {34.33} & {7.50} & {7.59} \\
        \midrule
        % NOTE:  must be applied per cell -- a macro argument cannot span "&"
        {\textbf{\OURS{} (Ours)}} & {\textbf{10.14}} & {\textbf{1.96}} & {\textbf{2.08}} \\
        \bottomrule
    \end{tabular}
\end{table}

\begin{table}[t]
    \centering
    \caption{{Quantitative evaluation of layout-conditioned scene generation. Both methods use randomly synthesized layouts. \OURS{} outperforms Seen2Scene in all metrics.}}
    \label{tab:layout_results}
    \small
    \setlength{\tabcolsep}{4pt}
    \begin{tabular}{lccc}
        \toprule
        Method & {FPD$\times 10^{2}$} ($\downarrow$) & {FID$\times 10^{-2}$} ($\downarrow$) & KID$\times 10^{2}$ ($\downarrow$) \\
        \midrule
        {Seen2Scene~\cite{meng2026seen2scene}} & {17.23} & {2.78} & {1.53} \\
        \midrule
        {\textbf{\OURS{} (Ours)}} & {\textbf{{11.79}}} & {\textbf{{2.21}}} & {\textbf{{1.44}}} \\
        \bottomrule
    \end{tabular}
\end{table}

\mypara{Synthetic scenes (3D-FRONT).}
\cref{tab:main_results} reports quantitative comparisons on the 3D-FRONT test split.
\OURS{} substantially outperforms all baselines in both geometric and perceptual metrics, {reducing FPD, FID, and KID by 59.6\%, 51.1\%, and 62.6\% compared to LT3SD.}
The FPD gain is likely due to our hierarchical generation, where a coarse {structure} and multi-resolution context guide global structural coherence and plausibility; 
{The FID/KID gains relate to} the high target resolution at {$1.5625\,\text{cm}$} voxel size, made tractable by our efficient sparse ROC tokenization.

\cref{fig:qualitative_comparison} confirms these improvements visually, showing that \OURS{} captures fine-scale structures such as thin chair legs and the hanging clothes in a closet, which baselines either smooth out or omit.
\cref{supp:3df_results} shows further comparisons on full rooms against OctGPT and LT3SD.

\begin{figure*}[htb]
  \centering
  \includegraphics[width=0.95\textwidth]{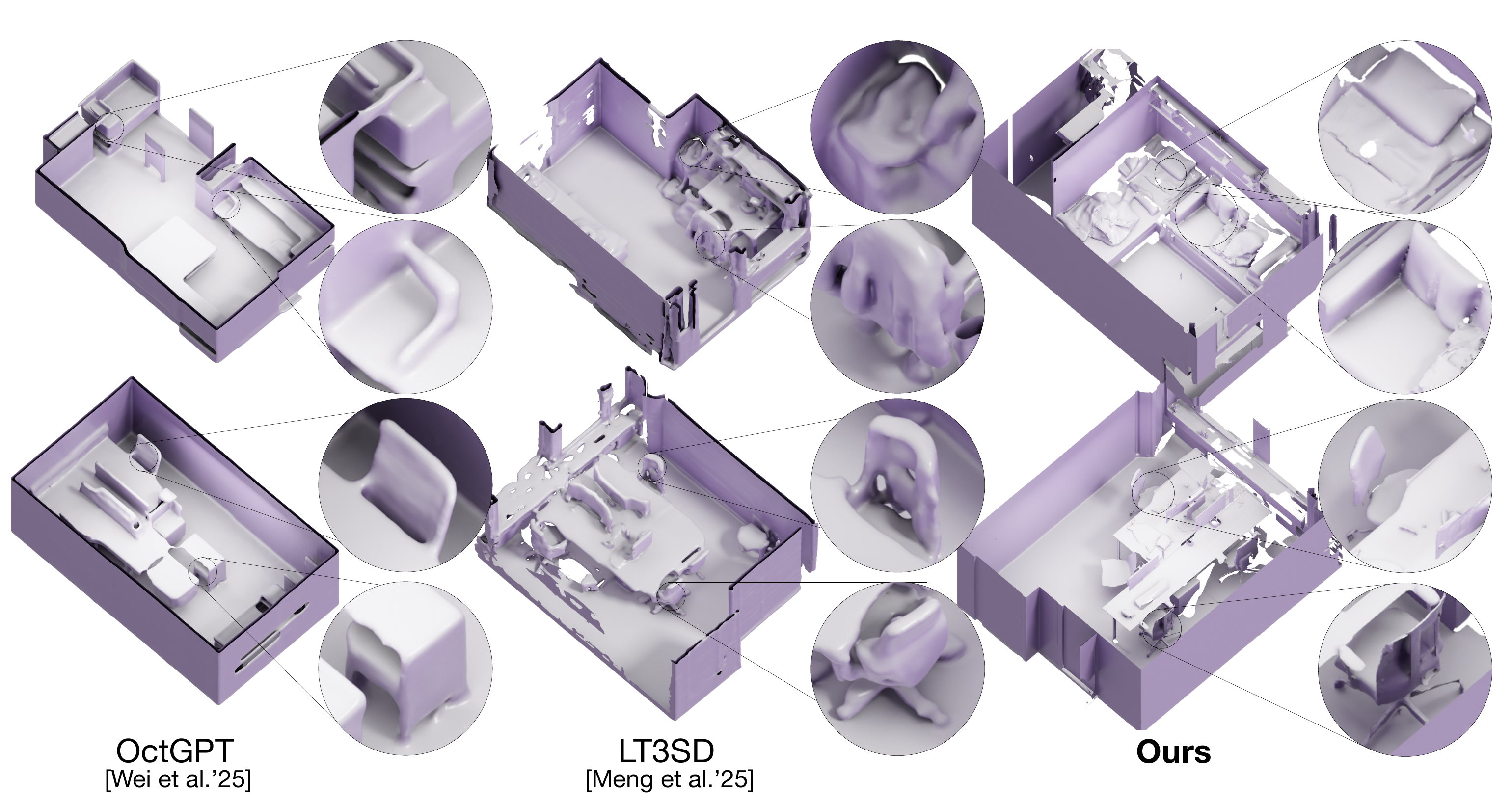}
  \par\vspace{-6pt}
  \caption{%
    Qualitative comparison on ScanNet++~\cite{yeshwanth2023scannet++}.
    \OURS{} generates fine geometric detail and realistic real-world clutter.
  }
  \label{fig:spp_scenes}
\end{figure*}

\mypara{Real-world scans (ScanNet++).}
\cref{tab:scannetpp_results} reports quantitative comparisons on the ScanNet++ test split.
{We compare with OctGPT and LT3SD, finetuning each released checkpoint on our ScanNet++ training split.}
\OURS{} again outperforms {both baselines} across both geometric and perceptual metrics, {reducing FPD, FID, and KID by 70.5\%, 73.8\%, and 72.5\% compared to LT3SD.}
The margins on the perceptual metrics widen further compared to 3D-FRONT, reflecting the increased realism of our generated scenes and their closer visual resemblance to real-world scans than the baselines.
\cref{fig:spp_scenes} shows the qualitative comparisons: \OURS{} reproduces realistic real-world clutter and fine geometric detail that {OctGPT and LT3SD smooth} into blob-like surfaces.
{\cref{supp:spp_results} shows more of our results.}

\begin{figure*}[hbt]
  \centering
  \includegraphics[width=\textwidth]{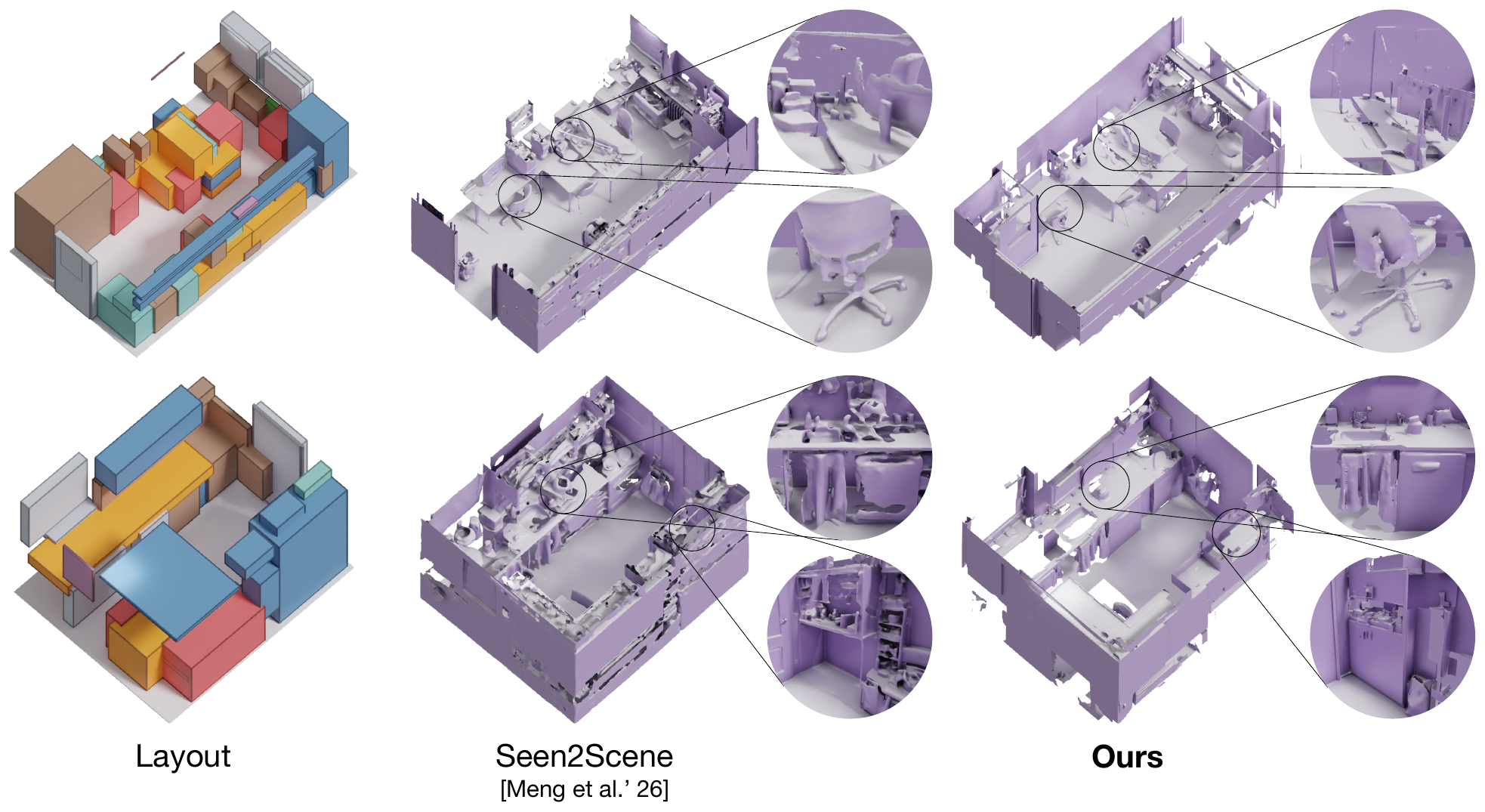}
  \par\vspace{-6pt}
  \caption{%
    Qualitative comparison of {layout-conditioned} scene generation. \OURS{} generates sharp surfaces and rich geometric details.
  }
  \label{fig:layout_generation}
\end{figure*}

\mypara{{Layout-conditioned generation.}}
To evaluate layout control beyond the layouts seen during training, we condition \OURS{} and Seen2Scene on randomly synthesized layouts that are not part of the training data,
and compare the generated scenes with ground-truth ScanNet++ chunks using the metrics above.
As shown in \cref{tab:layout_results}, \OURS{} reduces FPD, FID, and KID by 31.6\%, 20.7\%, and 6.1\% relative to Seen2Scene.
\cref{fig:layout_generation} compares \OURS{} with Seen2Scene under the same input layouts: \OURS{} follows the layouts with sharper surfaces and richer geometric detail.
 We refer to {\cref{supp:layout_results} for more results.}

\subsection{Ablations}

\begin{table}[t]
    \centering
    \caption{{Ablation study on 3D-FRONT. Incorporating sparse, multi-resolution ROC codes and using cascaded models enables our model to generate more diverse and realistic scenes.}}
    \label{tab:ablation}
    \small
    \setlength{\tabcolsep}{4pt}
    \begin{tabular}{lccc}
        \toprule
        Variant & \makecell{FPD\\$\times 10^{2}$} & \makecell{FID\\$\times 10^{-2}$} & \makecell{KID\\$\times 10^{2}$} \\
        \midrule
        {Single-resolution sequence} & {60.25} & {13.36} & {14.06} \\
        {Binary tokenization} & {29.61} & {7.40} & {4.87} \\
        {Dense formulation} & {24.01} & {5.19} & {2.60} \\
        {Unified model} & {26.38} & {5.63} & {2.75} \\
        \midrule
        {\textbf{\OURS{} (Ours)}} & {\textbf{11.03}} & {\textbf{2.40}} & {\textbf{1.49}} \\
        \bottomrule
    \end{tabular}
\end{table}

\mypara{Effect of ROC.}
We replace ROC with a naive 2-entry binary tokenizer that emits one occupancy bit per voxel, keeping the chunking, multi-resolution sequence, and transformer backbone identical, and extending the context length to 8{,}192 to fit the $8\times$ longer stream.
As shown in \cref{tab:ablation}, our ROC yields substantially lower FPD and FID at the same target resolution, indicating that ROC's per-token geometric context contributes substantially to quality gain beyond  multi-resolution architecture alone. 

\mypara{Do multi-resolution tokens help?}
We formulate {$\{\Phi^{(k)}\}$} at the $0.5\,\text{m}$ chunk extent with a single-resolution sequence containing only $3.125\,\text{cm}$ target tokens, and restricting cross-chunk conditioning to the same single resolution.
As shown in \cref{tab:ablation}, our multi-resolution sequencing yields substantially lower FPD and FID, indicating that progressive coarse-to-fine context provides anchoring that single-level sequencing cannot recover. 

\mypara{Effect of sparse formulation.}
We consider a dense formulation that retains every octree position (including the $T{=}0$ tokens our sparse formulation drops) across the multi-resolution context, the boundary prefix, and the GRU input.
As shown in \cref{tab:ablation}, our sparse formulation yields {lower FPD, FID, and KID, as well as} substantially faster inference, since the per-chunk sequence length scales with the occupied geometry rather than the full octree volume. For a comprehensive comparison of the sequence length and inference speed, we refer to {\cref{supp:sparse_vs_dense}}.

\mypara{Can one model replace the cascade?}
We consider a unified transformer trained jointly on chunks at all four cascade extents -- enabled by ROC's resolution-invariance and our resolution-agnostic sequencing -- in place of the four cascaded models.
As shown in \cref{tab:ablation}, the unified model underperforms the cascade because the per-chunk sequence distribution differs substantially across resolutions, and a single model cannot fit all four regimes simultaneously. 

\mypara{Limitations.}
\OURS{} is designed for single-floor indoor scenes; multi-story buildings would require a height-aware positional encoding that models discrete floors. The cascade comprises $K{+}1$ independently trained transformers, so generating at a finer voxel resolution requires training an additional stage rather than extending an existing one.
On real-world data, our model inherits the holes and noise of the training scans; learning to complete partial geometry at scene scale is a natural direction for future work.

\section{Conclusion}
\label{sec:conclusion}
We introduced \OURS{}, a sparse autoregressive approach for high-fidelity 3D indoor scene generation in a coarse-to-fine fashion. 
We tokenize voxel octants with a training-free, resolution-invariant codebook and serialize sparse multi-resolution scene hierarchies, casting scene synthesis as chunk-by-chunk super-resolution conditioned on neighbor boundaries. 
This enables synthesizing scenes with sharp surfaces and rich geometric details at finer voxel resolutions than prior work, on both synthetic and real-world data. 
We believe \OURS{} will drive progress in realistic 3D scene generation, opening new possibilities for embodied agents grounded in real-world indoor geometry.

%%%%%%%%% ACKNOWLEDGMENTS - comment out for the anonymous review version
\section*{Acknowledgments}
This project is funded by the ERC Starting Grant SpatialSem (101076253). We also acknowledge the EuroHPC Joint Undertaking for providing access to the EuroHPC supercomputers MareNostrum 5, hosted by the Barcelona Supercomputing Center (BSC), Spain, and LEONARDO, hosted by CINECA, Italy.

{
    \small
    \bibliographystyle{ieeenat_fullname}
    \bibliography{main}
}

% WARNING: do not forget to delete the supplementary pages from your submission
\clearpage
\setcounter{page}{1}
\maketitlesupplementary
% Appendix sections are numbered A, B, C, ...
\appendix
% Supplementary material. Included from sec/X_suppl.tex after \appendix,
% so the sections below are numbered A, B, C, ...
{We present implementation details,} additional ablations and a detailed runtime / sparsity comparison, details of layout-guided generation and mesh extraction, and additional 3D-FRONT comparisons, ScanNet++ results, and layout-conditioned results.

\section{{Implementation Details}}
\label{supp:impl}
\mypara{{Data processing.}}
{For each room, we voxelize the geometry at a leaf size of $1.5625\,\text{cm}$ and aggregate up to $2\,\text{m}$ via $2{\times}2{\times}2$ max-pooling, producing the multi-resolution input to the coarse and cascade stages.
Each occupied $2\,\text{m}$ voxel seeds a $2\,\text{m}{\times}2\,\text{m}{\times}2\,\text{m}$ chunk for the first cascade transformer; subsequent stages subdivide these chunks laterally as described in \cref{para:chunk_refinement}.
Rooms are bounded by a $10\,\text{m}{\times}10\,\text{m}$ footprint and cropped at a height of $2\,\text{m}$, so that every chunk spans the full scene height.}

\mypara{{Training.}}
{We implement \OURS{} in PyTorch and train all transformers with the AdamW optimizer at a peak learning rate of $1{\times}10^{-4}$.
The coarse scene transformer $\Phi^{(0)}$ and the $K{=}4$ cascade transformers $\{\Phi^{(k)}\}_{k=1}^{4}$ share a GPT-2 decoder backbone with 24 layers, a hidden dimension of 1024, and 16 attention heads.
The cascade context length is 4{,}800 tokens for the three coarser stages ($2\,\text{m}$, $1\,\text{m}$, and $0.5\,\text{m}$ chunks) and 8{,}192 tokens for the finest stage ($0.25\,\text{m}$ chunks), whose sparse sequences are longer.
The per-GPU batch size scales inversely with chunk-token length, ranging from 16 to 4.
Each transformer trains for $\approx 4$ days on four NVIDIA H100 GPUs.
The layout conditioning and the mesh VAE $\Phi_m$ are detailed in \cref{supp:layout_guided,supp:mesh}.}

\section{{Evaluation Protocol}}
\label{supp:eval_protocol}
We evaluate all methods with {1000} sampled scene chunks, comparing generated and ground-truth meshes in FPD, FID, and KID. For FPD, we sample surface points per chunk, normalized to the unit sphere, and encode {them with the pretrained Uni3D encoder~\cite{zhou2024uni3d}, $\ell_2$-normalizing the features before computing the Fr\'echet distance to the ground truth}. For FID and KID, we render each chunk from 8 viewpoints uniformly spaced around the vertical axis under fixed lighting{, \ie, 8{,}000 images per method, and use the CLS token of DINOv2 ViT-B/14~\cite{oquab2023dinov2} as the image feature}.
{For the ablations (\cref{tab:ablation}), each variant refines a subset of the coarse scenes of our full model. Since FPD, FID, and KID all depend on the number of samples, we compare each variant with our full model on the chunks at the same locations of the same scenes, using all of them up to 1000.}

\section{{User Study}}
\label{supp:user_study}
We conducted a user study with 22 participants on three aspects: perceptual quality, room-type consistency, and layout controllability. We will expand on this study in the revision.
(1) Perceptual quality: \OURS{} achieved 84.6\% overall preference vs baselines (86.2\% geometric fidelity, 76.7\% realism, 75.4\% layout coherence).
(2) Room type consistency: given our generated scenes, participants correctly identified the input room label with 90.6\% accuracy.
(3) Layout controllability: under the finer-grained control section described below, participants rated scene-layout alignment 4.37/5.

\section{{Runtime}}
\label{supp:runtime}
{On a single NVIDIA H100 GPU, \OURS{} generates a room in $22$ minutes on average. The cascade super-resolution stage dominates at $\sim\!20$ minutes; the coarse predictor and mesh extraction take $\sim\!1$ minute and $\sim\!30$ seconds, respectively.
For a $4\,\text{m}\times4\,\text{m}$ scene, \OURS{} takes $16$ minutes as we generate at 41\% finer resolution ($1.5625\,\text{cm}$), compared to LT3SD~\cite{meng2025lt3sd} ($8$ minutes) and WorldGrow~\cite{li2026worldgrow} ($6$ minutes). At an output resolution of $3.125\,\text{cm}$, closer to that of the baselines, \OURS{} generates in $3.5$ minutes.
Our implementation is not optimized for speed; standard transformer-decoding optimizations (KV cache, batched decoding, speculative sampling) could accelerate generation.
\Cref{tab:sparse_ablation} breaks down the generation time per cascade level and compares it with the dense formulation (\cref{supp:sparse_vs_dense}).}

\begin{table}[t]
    \centering
    \caption{Ablation: sparse vs.\ dense formulation. {Generation time per room and mean number of tokens per chunk at each cascade level $k$.}}
    \label{tab:sparse_ablation}
    \small
    \setlength{\tabcolsep}{4pt}
    \begin{tabular}{lccccc}
        \toprule
         & \multicolumn{3}{c}{Gen.\ time (s)} & \multicolumn{2}{c}{Tokens / chunk} \\
        \cmidrule(lr){2-4} \cmidrule(lr){5-6}
        Level & Dense & Sparse & Speedup & Dense & Sparse \\
        \midrule
        1          & 36.0     & 10.4    & 3.5$\times$  & 717   & 251 \\
        2          & 234.4    & 49.6    & 4.7$\times$  & 1{,}495 & 383 \\
        3          & 1{,}988.0 & 218.5  & 9.1$\times$  & 3{,}059 & 502 \\
        4          & 24{,}999.0 & 966.2 & 25.9$\times$ & 6{,}195 & 621 \\
        \midrule
        Room (total) & 27{,}334.5 & \textbf{1{,}320.5} & \textbf{20.7$\times$} & --- & --- \\
        \bottomrule
    \end{tabular}
\end{table}

\section{{Layout-guided Generation}}
\label{supp:layout_guided}

\mypara{{Layout tokens.}}
{The Fourier features $\gamma$ use eight frequency bands, $\sin(2^j\pi x)$ and $\cos(2^j\pi x)$ for $j = 0, \dots, 7$, applied to each normalized box coordinate $x$.
The layout tokens enter the transformer through a zero-initialized linear gate, so that training starts from the unconditional model.}

\mypara{{Semantic field.}}
{The field of $\Phi^{(0)}$ is defined on the $20 \times 20 \times 4$ grid of $50\,\text{cm}$ cells covering the $10\,\text{m} \times 10\,\text{m} \times 2\,\text{m}$ scene bound.
We rasterize the boxes at $25\,\text{cm}$ and pool them to the coarse cells.
Each cell stores the overlap-weighted mean of the $\ell_2$-normalized CLIP features of the boxes covering it, together with three coverage values: the covered fraction, the share of the dominant box, and the coverage of its $25\,\text{cm}$ child cells.
The field is projected and added onto the coarse tokens through a zero-initialized gate, and each coarse token additionally receives the box of its own cell through the same Fourier encoder.}

\mypara{{Box selection.}}
{Both stages use at most $96$ layout tokens, which only affects rooms with more than $96$ boxes.
$\Phi^{(0)}$ keeps the boxes that overlap the most coarse cells, and $\Phi^{(k)}$ keeps the boxes whose centers are nearest to the center of chunk $C$.
The resulting sequence of $\Phi^{(k)}$ is $[\mathcal{H}_C, \mathcal{B}_{\text{global}}, \mathcal{B}_{\text{local}}, \mathcal{S}_C^{(k)}]$, between the start and end tokens.}

\section{{Mesh Extraction}}
\label{supp:mesh}
{We detail the architecture, training, and inference of the sparse VAE $\Phi_m$ (\cref{subsec:mesh}).}

\mypara{{Architecture.}}
{$\Phi_m$ follows the dual-grid SC-VAE of TRELLIS.2~\cite{xiang2025native} and is initialized from its pretrained weights.
The encoder receives only the occupied voxels of $O^{(K)}$ and downsamples them in four stages of sparse ConvNeXt blocks (64 to 1024 channels) into a 32-channel sparse latent with $25\,\text{cm}$ cells.
The decoder mirrors the encoder: each of its four upsampling blocks predicts the occupancy logits of the eight children of every voxel and keeps the children with positive logits, and the finest level predicts the dual-vertex offsets, edge-intersection flags, and quad-lerp weights.
In the original SC-VAE, the upsampling blocks copy the child structure cached by the encoder, so the output voxel set always equals the input; we remove this cache, so that the predicted child occupancy determines the output structure.
In addition, the latent voxel set is dilated by its 26-neighbors before decoding, which allows the decoder to add geometry in empty $25\,\text{cm}$ cells next to the input.}

\mypara{{Training data.}}
{The ground-truth dual-grid attributes of each room are computed from its mesh at $v_K = 1.5625\,\text{cm}$, keeping the geometry below $2\,\text{m}$.
We train on 3D-FRONT~\cite{fu20213d}, ScanNet++~\cite{yeshwanth2023scannet++}, and ARKitScenes~\cite{baruch2021arkitscenes} and translate each room by a random integer offset of 0--15 voxels.}

\mypara{{Input augmentation.}}
{For 95\% of the training samples, we corrupt the encoder input with artifacts modeled on those of the generated occupancy, while the decoder is supervised with the clean ground truth:
(1) small surface defects at 3--8 per 1{,}000 occupied voxels, in clusters of 2--6 within a radius of 5 voxels, half of which remove runs of 1--3 voxels through the surface and half of which add stray voxels or patches of 3--8 voxels on top of it, with 60\% of the clusters placed where oblique or curved surfaces step between voxel layers;
(2) irregular holes of 20--600 voxels grown along the surface, removing 1--5\% of the voxels;
(3) random removal of up to 0.5\% of the voxels;
(4) spurious voxels adjacent to the surface, up to 3\% of the occupied voxels; and
(5) up to two floating blobs of 2--4 voxels.}

\mypara{{Losses.}}
{During training, each upsampling block continues with the union of its predicted and ground-truth children, so that all ground-truth voxels receive supervision.
The objective combines a binary cross-entropy on the child occupancy of every upsampling level (weight 0.2), a binary cross-entropy on the edge-intersection flags (0.1), a mean squared error on the dual-vertex offsets (0.01), a KL term ($10^{-6}$), and the rendering losses of TRELLIS.2 on the extracted mesh: L1 losses on the rendered mask (1) and depth (10), and L1 (1), SSIM~\cite{wang2004image} (0.2), and LPIPS~\cite{zhang2018lpips} (0.2) losses on the rendered normals.
Each step renders one $512 \times 512$ view per room from a camera placed in the free space of the room, looking at a random occupied voxel with a $70^\circ$ field of view.}

\mypara{{Optimization.}}
{Starting from the pretrained weights, $\Phi_m$ is trained in several stages.
The final stage runs for $30{,}000$ iterations with AdamW at a learning rate of $10^{-5}$ (500 warm-up iterations, cosine decay) and a batch size of 4 on two NVIDIA RTX A6000 GPUs, and we use its EMA weights (rate 0.999).}

\mypara{{Inference.}}
{At inference, each upsampling block keeps only its predicted children.
We lower the decision threshold of the edge-intersection flags from a logit of 0 to $-3$, which reduces pinholes on flat surfaces.
Finally, we close small holes in the extracted mesh: boundary loops of at most 6 edges are fan-filled, and holes with rims of at most 64 edges and up to $25\,\text{cm}$ across are closed with minimum-area patches.}

\section{Sparse vs.\ Dense Formulation}
\label{supp:sparse_vs_dense}
{\Cref{tab:sparse_ablation}} reports per-level generation time and mean sequence length, comparing our sparse formulation against the dense baseline.
The speedup grows with refinement depth---$3.5\times$ at level $1$ to $25.9\times$ at level $4$---driven by sequence length: a level-$4$ dense chunk averages $6{,}195$ tokens versus $621$ for sparse, and transformer decoding is super-linear in length.
End-to-end, sparse generation takes $1{,}321\,\text{s}$ per room versus $27{,}335\,\text{s}$ for dense, a $20.7\times$ overall speedup, with every one of the $33$ rooms speeding up by at least $13\times$.
The dense baseline is in fact prohibitively slow at level $4$, completing only $38$ of the $101$ test rooms within the same compute budget; {the} sparse formulation is therefore a prerequisite for scaling to fine resolutions.

\section{{Additional Comparisons on 3D-FRONT}}
\label{supp:3df_results}
\begin{figure*}[t]
  \centering
  \includegraphics[width=\textwidth]{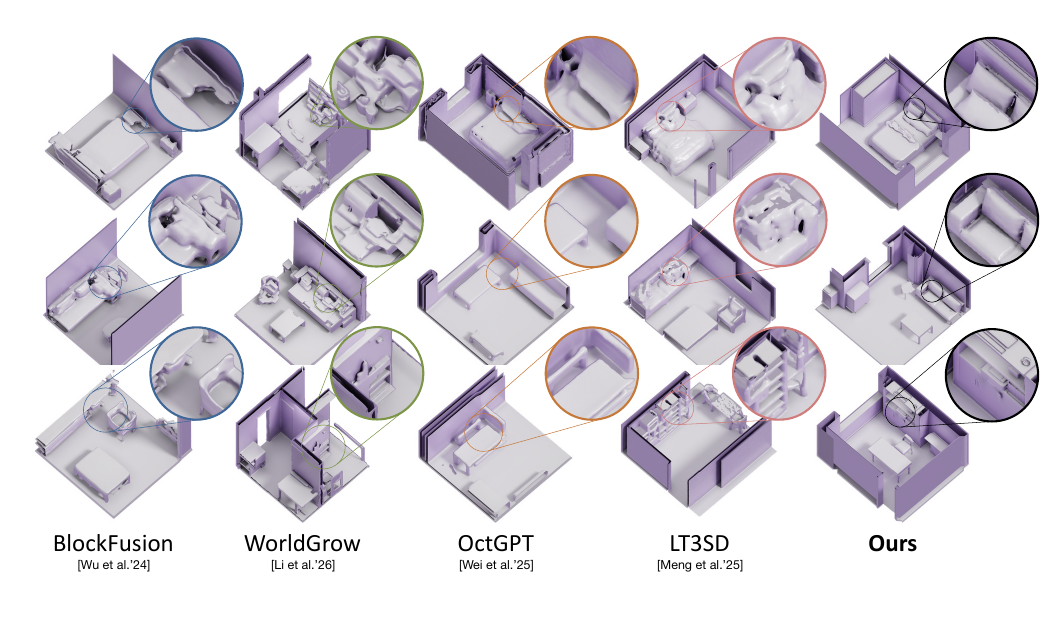}
  \caption{%
    {Additional qualitative comparison on 3D-FRONT. \OURS{} generates sharp surfaces and rich geometric details than the baselines.}
  }
  \label{fig:qualitative_comparison_more}
\end{figure*}

\begin{figure*}[htb]
  \centering
  \includegraphics[width=0.95\textwidth]{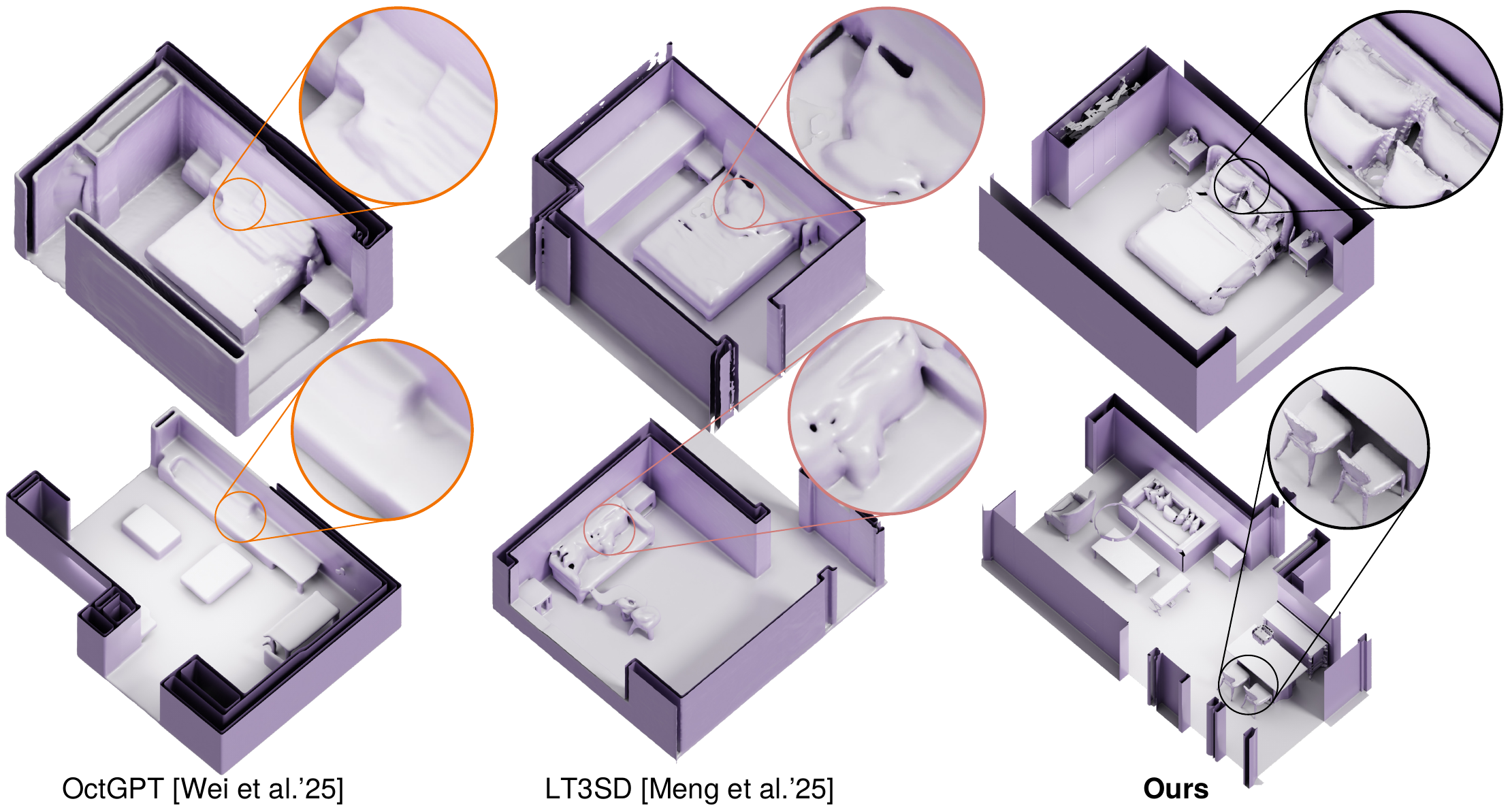}
  \caption{%
    Qualitative comparison on 3D-FRONT~\cite{fu20213d}.
    \OURS{} recovers finer geometric detail and more coherent room-scale layouts.
  }
  \label{fig:3df_scenes}
\end{figure*}

{\cref{fig:qualitative_comparison_more} shows further comparisons with the baselines on 3D-FRONT, as in \cref{fig:qualitative_comparison}.}
{\cref{fig:3df_scenes} compares full scenes generated by \OURS{}, OctGPT, and LT3SD.}

% moved from the main paper (sec/4_experiments.tex) to save space

\section{{Additional Results on ScanNet++}}
\label{supp:spp_results}
\begin{figure*}[htb]
  \centering
  \includegraphics[width=0.95\textwidth]{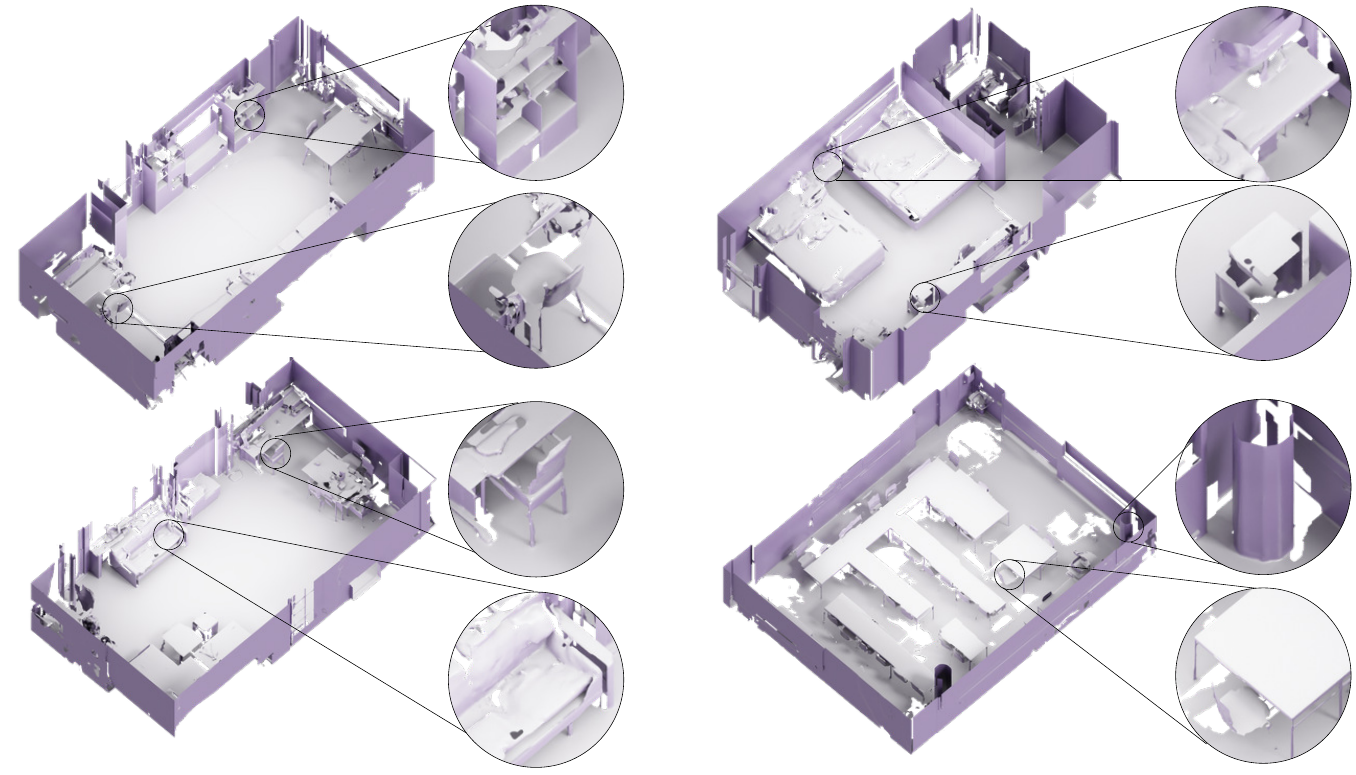}
  \caption{%
    3D scene generation from the real scan distribution of ScanNet++, showing both fine geometric structures and detail along with coherent room-scale structures.
  }
  \label{fig:spp_ours_main}
\end{figure*}

{\cref{fig:spp_ours_main} shows additional \OURS{} generations from the ScanNet++ distribution.}

\section{{Additional Layout-conditioned Results}}
\label{supp:layout_results}
\begin{figure*}[htb]
  \centering
  \IfFileExists{assets/layout_ours_main.pdf}{%
    \includegraphics[width=0.95\textwidth]{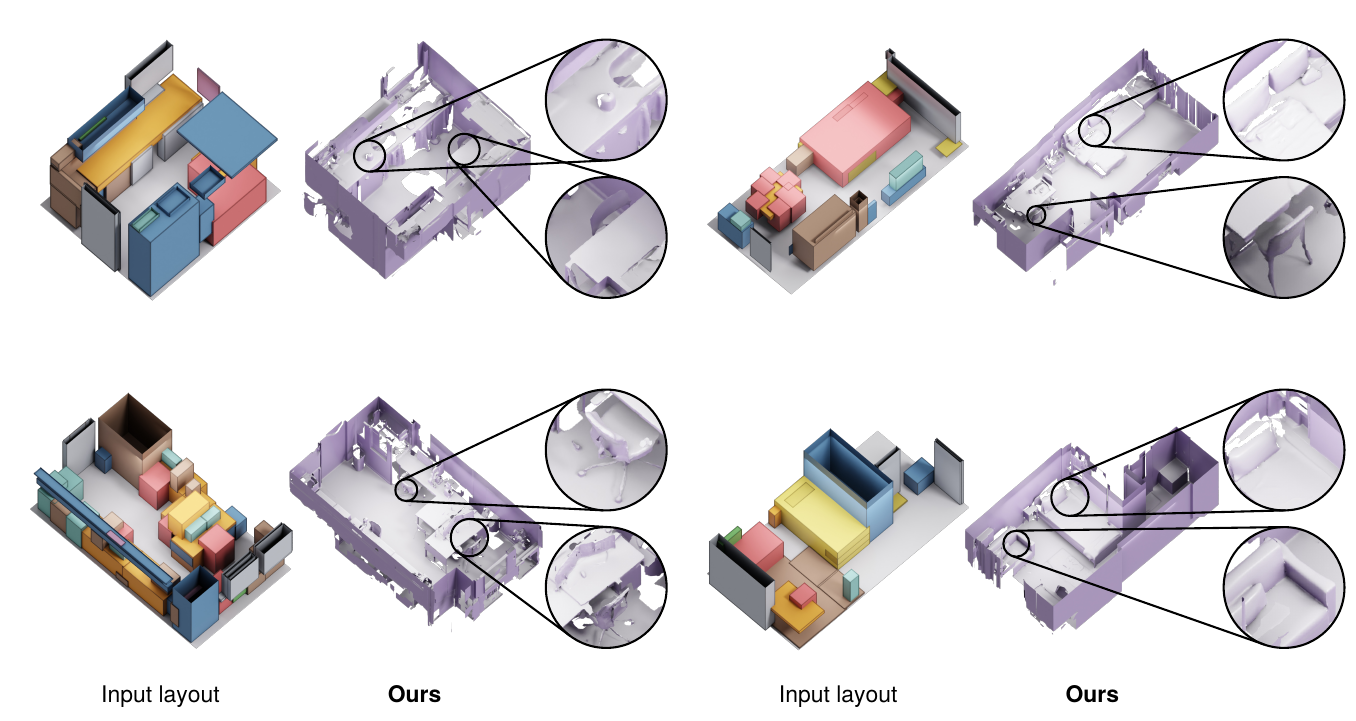}%
  }{%
    \fbox{\parbox[c][5cm][c]{0.9\textwidth}{\centering Layout-conditioned results of \OURS{} (pending: scenes and close-ups are being selected).}}%
  }
  \caption{%
    {Layout-conditioned scene generation on ScanNet++. \OURS{} generates sharp surfaces capturing the real-world distributions.}
  }
  \label{fig:layout_ours_main}
\end{figure*}

{\cref{fig:layout_ours_main} shows additional scenes generated by \OURS{} from the object layouts of real ScanNet++ rooms, as in \cref{fig:layout_generation}.}

\end{document}